\documentclass[sigconf]{acmart}

\copyrightyear{2020} 
\acmYear{2020} 
\setcopyright{acmlicensed}\acmConference[PPMLP'20]{2020 Workshop on Privacy-Preserving Machine Learning in Practice}{November 9, 2020}{Virtual Event, USA}
\acmConference{2020 Workshop on Privacy-Preserving Machine Learning in Practice (PPMLP'20), November 9, 2020, Virtual Event, USA}
\acmPrice{15.00}
\acmDOI{10.1145/3411501.3419419}
\acmISBN{978-1-4503-8088-1/20/11}

\usepackage{microtype}
\usepackage{graphicx}
\usepackage{subcaption}
\usepackage{booktabs} % for professional tables

\usepackage{hyperref}

\AtBeginDocument{%
  \providecommand\BibTeX{{%
    \normalfont B\kern-0.5em{\scshape i\kern-0.25em b}\kern-0.8em\TeX}}}

\begin{document}
\fancyhead{}
%%
%% The "title" command has an optional parameter,
%% allowing the author to define a "short title" to be used in page headers.

%% Not Private & Not Fair: Impact of Data Imbalance on Utility and Fairness in Differential Privacy
%% Neither Private Nor Fair: Impact of Data Imbalance in Differential Privacy
\title{Neither Private Nor Fair: Impact of Data Imbalance on Utility and Fairness in Differential Privacy}

%%
%% The "author" command and its associated commands are used to define
%% the authors and their affiliations.
%% Of note is the shared affiliation of the first two authors, and the
%% "authornote" and "authornotemark" commands
%% used to denote shared contribution to the research.
\author{Tom Farrand}
\email{tom@seldon.io}
\affiliation{%
  \institution{Seldon}
  \institution{OpenMined}
}

\author{Fatemehsadat Mireshghallah}
\email{fmireshg@eng.ucsd.edu}
\affiliation{%
  \institution{University of California San Diego}
  \institution{OpenMined}
}

\author{Sahib Singh}
\email{sahibsin@alumni.cmu.edu}
\affiliation{%
  \institution{Ford R\&A}
  \institution{OpenMined}
}

\author{Andrew Trask}
\email{andrew@openmined.org}
\affiliation{%
  \institution{University of Oxford}
  \institution{OpenMined}
}

%%
%% By default, the full list of authors will be used in the page
%% headers. Often, this list is too long, and will overlap
%% other information printed in the page headers. This command allows
%% the author to define a more concise list
%% of authors' names for this purpose.
\renewcommand{\shortauthors}{Farrand and Mireshghallah, et al.}

%%
%% The abstract is a short summary of the work to be presented in the
%% article.
\begin{abstract}
Deployment of deep learning in different fields and industries is growing day by day due to its performance, which relies on the availability of data and compute. Data is often crowd-sourced and contains sensitive information about its contributors, which leaks into models that are trained on it. To achieve rigorous privacy guarantees, differentially private training mechanisms are used.  However, it has recently been shown that differential privacy can exacerbate existing biases in the data and have disparate impacts on the accuracy of different subgroups of data.  In this paper, we aim to study these effects within differentially private deep learning. Specifically, we aim to study how different levels of imbalance in the data affect the accuracy and the fairness of the decisions made by the model, given different levels of privacy. We demonstrate that even small imbalances and loose privacy guarantees can cause disparate impacts. 
\end{abstract}

%%
%% Keywords. The author(s) should pick words that accurately describe
%% the work being presented. Separate the keywords with commas.
\begin{CCSXML}
<ccs2012>
<concept>
<concept_id>10002978.10003029.10003032</concept_id>
<concept_desc>Security and privacy~Social aspects of security and privacy</concept_desc>
<concept_significance>500</concept_significance>
</concept>
<concept>
<concept_id>10002978.10003029.10011150</concept_id>
<concept_desc>Security and privacy~Privacy protections</concept_desc>
<concept_significance>500</concept_significance>
</concept>
<concept>
<concept_id>10002978.10003029.10011703</concept_id>
<concept_desc>Security and privacy~Usability in security and privacy</concept_desc>
<concept_significance>500</concept_significance>
</concept>
<concept>
<concept_id>10002978.10002991.10002995</concept_id>
<concept_desc>Security and privacy~Privacy-preserving protocols</concept_desc>
<concept_significance>300</concept_significance>
</concept>
<concept>
<concept_id>10010147.10010257.10010293.10010294</concept_id>
<concept_desc>Computing methodologies~Neural networks</concept_desc>
<concept_significance>500</concept_significance>
</concept>
</ccs2012>
\end{CCSXML}

\ccsdesc[500]{Security and privacy~Social aspects of security and privacy}
\ccsdesc[500]{Security and privacy~Privacy protections}
\ccsdesc[500]{Security and privacy~Usability in security and privacy}
\ccsdesc[300]{Security and privacy~Privacy-preserving protocols}
\ccsdesc[500]{Computing methodologies~Neural networks}

\keywords{Fairness, Bias, Differential Privacy, Data Imbalance, Deep Learning}
%%
%% This command processes the author and affiliation and title
%% information and builds the first part of the formatted document.
\maketitle

\section{Introduction}

Given the high performance of deep learning algorithms in a variety of tasks, they are widely deployed. These algorithms rely on large data sets and high performance compute to perform well.
This data is often crowd-sourced and likely contains sensitive information about its contributors~ \cite{nytimes2}. %
It has been shown widely in the literature that machine learning models, and more specifically DNNs, tend to memorize information from the training set~ \cite{property19, memorize, Mireshghallah2020PrivacyID}.
There are a plethora of attacks that exploit the vulnerabilities in DNNs and extract sensitive information (e.g. gender, ethnicity, genetic markers) from models in both black and white box settings ~\cite{shokri, pharmacogenetics}. 

To mitigate this, Differential Privacy (DP) is used.  DP can protect against attempts to infer the contribution of a given individual to the training set by adding noise to the computations~\cite{chaudhuriERM, abadi2016deep}. DP is used in many contexts, including in medical settings \cite{singh2020benchmarking} and is also  used to protect 2020 US Census data \cite{abowd2018us}. 
The most prominent DP algorithm in deep learning is DP-Stochastic Gradient Descent (DP-SGD)~\cite{abadi2016deep} wherein noise is added to clipped gradients during training. In this work, ~\citeauthor{abadi2016deep} introduce the moments accountant, which is a method that keeps track of the exhausted privacy budget over time and amplifies privacy during training. The moments accountant returns the overall budget used and the privacy leakage, which are demonstrated with $\epsilon$ and $\delta$, respectively. 
Although DP-SGD offers rigorous privacy guarantees, it degrades the accuracy of the resulting model. Furthermore, it has recently been shown that this degradation is disparate; minority subgroups of data suffer more utility loss compared to others~\cite{bagdasaryan2019differential, kuppam2019fair}. 

~\citeauthor{bagdasaryan2019differential}~\cite{bagdasaryan2019differential} empirically show that if DP-SGD is used on data which is highly imbalanced, as in contains subgroups with extremely small populations, the less represented groups which already have lower accuracy end up losing more utility: "the poor become poorer".
They also show that as stricter privacy guarantees are imposed, this gap gets wider.
This gap can have huge societal and economic implications. For instance, ~\cite{kuppam2019fair} shows that if DP was used in the decision making of a fund allocation problem (based on US Census data), smaller districts would end up with comparatively more funding than what they would receive without DP, and larger districts would get less funding. 

To achieve a deeper insight into the effects of differential privacy, we set out to study in detail the behavior of deep learning models trained using DP-SGD. 
To be more precise, we set out to study a wider range of imbalance than what was explored by~\citeauthor{bagdasaryan2019differential}. They studied datasets in which the minority subgroup formed $0.01\%-5.00\%$ of the entire data.
We, on the other hand, study less significant imbalances as well. We cover a range of $0.1\%-30\%$ for the population of the minority subgroup ($30\%$ might not be considered a minority, however it is a common imbalance to have in datasets).
We also explore a wider range of privacy budget. ~\citeauthor{bagdasaryan2019differential} study $\epsilon$ in range of $3$ to $10$, whereas we also consider $\epsilon = 1.15$ and $\epsilon=16.2$.
Our results show that if there are two subgroups in the data and the minority group comprises $30\%$ the data (CelebA data and the minority group is male), which is actually a large portion and is the case for many datasets, using differential privacy with $\epsilon=16.2$ and $\epsilon=4.98$ yields $5.1\%$ and $5.2\%$ more disparity in accuracy between the two subgroups (female vs. male). We show that increasing the imbalance (i.e. decreasing the population of the minority subgroup to $0.1\%$) increases this disparity to $19\%$ and $52\%$ respectively. This result shows that even when differential privacy is applied with loose guarantees, on datasets with negligible imbalance, it can have a huge impact on the minority subgroups. We also study the fairness metrics demographic parity and equality of opportunity, finding that as datasets become more imbalanced and stricter privacy guarantees are used the fairness, with respect to these two metrics, of models is reduced. %CHECK WITH TOM! 
%

%
%More specifically, we have selected the CelebA dataset (a dataset different from those used in~\cite{bagdasaryan2019differential}), and evaluated a ResNet18's ability to perform smile detection under varying DP conditions.
%
%We experiment with both data imbalance ratios, as well as with the hyperparameter settings used by DP-SGD- specifically the amount of noise applied ($\sigma$) and the gradient clipping level ($S$), leading to a consideration of three different differential privacy levels.

\section{Related Work}
%Does mitigating ML’s impact disparity require treatment disparity?
%Fair Decision Making Using Privacy-Protected Data Generative Adversarial Models
%Differential Privacy Has Disparate Impact on Model Accuracy
%batch citation: Actionable Recourse in Linear Classification, Censored and Fair Universal Representations using Generative Adversarial Models, Differentially Private Fair Learning
%
While there have been several papers related to fairness and differential privacy, works most relevant to ours are mentioned below.

\citet{bagdasaryan2019differential} empirically demonstrates a disproportionately large reduction in accuracy for subgroups when modeled using neural networks trained with DP-SGD. For example, a gender classification model trained using DP-SGD on the Flickr-based Diversity in Faces (DiF) dataset \cite{merler2019diversity} and the UTKFace dataset \cite{zhang2017age} dataset has a much lower accuracy when classifying black faces compared to white faces. Similar results were also found when performing sentiment analysis on a corpus of African-American English tweets \cite{blodgett2016demographic,blodgett2018twitter}, species classification on iNaturalist nature images \cite{van2018inaturalist}, and federated learning of a language model on a public Reddit comments dataset \cite{mcmahan2016communication}. As discussed in the introduction, our work is different from the one by~\citeauthor{bagdasaryan2019differential} in that we study a wider range of privacy budgets and also a wider range of data imbalances. Apart from that, we also use other notions of privacy, such as equality of opportunity and demographic parity. 

The work by \citet{kuppam2019fair} proposes novel measures of fairness in the context of randomized DP algorithms and identifies causes of outcome disparities. 

Our paper builds on these above works by experimenting with varied data imbalance ratios and seeks to measure the difference in equality of opportunity and demographic parity, common metrics of fairness.
Other seminal papers addressing fair decision making include \cite{lipton2018does, ustun2019actionable, kairouz2019censored, jagielski2018differentially}

\section{Background}
In this section, we discuss the fundamental privacy and fairness concepts used throughout the paper.

\subsection{Differentially Private SGD (DP-SGD)}
Differential Privacy \cite{dwork2006our,dwork2006calibrating} provides a strong privacy guarantee for algorithms on aggregate databases. 
For  $\epsilon  \geq 0$, an algorithm $A$ is understood to satisfy Differential Privacy if and only if for any pair of datasets that differ in only one element, the following statement holds true.
\begin{center}
$P[A(D) = t] \leq e^\epsilon$ $P[A(D^{'}) = t] $ $\; \forall t$
\end{center}
Where $D$ and $D'$ are differing datasets by at most one element, and $P[A(D) = t]$ denotes the probability that $t$ is the output of $A$. This setting approximates the effect of individual opt-outs by minimising the inclusion effect of an individual’s data. 

Stochastic Gradient Descent (SGD) is an iterative method for optimising differentiable objective functions. It updates weights and biases by calculating the gradient of a loss function on small batches of data.
DP-SGD \cite{abadi2016deep} is a modification of the stochastic gradient descent algorithm which provides provable privacy guarantees. It is different from SGD because it bounds the sensitivity of each gradient and is paired with a moments accountant algorithm to amplify and track the privacy loss across weight updates. The moments accountant accumulates and tracks the privacy expenditure in the training process of deep neural networks. Moments accountant significantly improves on earlier privacy analysis of SGD and allows for meaningful privacy guarantees for deep learning trained on realistically sized datasets \cite{bu2019deep}.

In order to ensure SGD is differentially private (i.e. DP-SGD), there are two modifications to be made to the original SGD algorithm. First, the sensitivity of each gradient must be bounded. This is done by clipping the gradient in the $l2$ norm. Second, one applies random noise to the earlier clipped gradient, multiplying its sum by the learning rate, and then using it to update the model parameters.

\subsection{Fairness \& Bias In Machine Learning}
Fairness in machine learning is a measure of the degree to which there is a disparate treatment for different groups which should have been treated equally (e.g. female vs. male). Hence an algorithm whose decisions skew towards a particular group of people can be said to be unfair.

The definitions of fairness can fall under individual fairness, group fairness and subgroup fairness \cite{mehrabi2019survey}. Individual fairness is where similar predictions are provided for similar individuals. Group fairness refers to equal treatment of various groups. Subgroup fairness obtains the best properties of the former groups by picking a group fairness constraint and asking if it holds over a large collection of subgroups \cite{kearns2019empirical}. 

Like people, algorithms are also susceptible to biases.
There are many different types of bias as discussed in \citet{mehrabi2019survey}. Historical Bias refers to the bias which already exists due to prior historical biases or socio-technical issues in society. For example, a 2018 image search result involving female CEOs showed fewer images since only 5\% of Fortune 500 CEOs were women. Representation Bias can result from the way we define and sample a population (e.g. lack of geographical diversity in ImageNet). Algorithmic bias is caused when a bias is generated by the algorithm without being present in the input data.

%\vspace{-ex}
\section{Experimental Results}
Our experiments implement a ResNet18 model \cite{resnet18}, pre-trained on ImageNet, to perform smile classification on a subset of 30,000 images from the CelebA dataset \cite{celebadataset}. During the course of training, we experiment with both data imbalance ratios, as well as with the hyperparameter settings used by DP-SGD- specifically the amount of noise applied ($\sigma$) and the gradient clipping level ($S$). Following \citet{bagdasaryan2019differential} we compute privacy budget $\epsilon$ for each training run using the Rényi DP \cite{mironov17} implementation. 

We implement our experiments using PyTorch \cite{pytorchlibrary} and ran them on four Nvidia K80 GPUs. After hyperparameter optimisation using grid search; we use a batch size of 32, a learning rate of 0.00005, and train for 60 epochs. Adam optimiser is used to minimise our loss function. A full training run of 60 epochs with differential privacy applied takes 20 hours to converge in this setup. 

The gender balance of the CelebA subset was manipulated to ensure that females were always the majority. The gender imbalances considered were: 99.9\% (29,970) female to 0.1\% male (30), 99\% (29,700) to 1\% (300) male, 90\% (27,000) female to 10\% (3,000) male, 80\% (24,000) female to 20\% (6,000) male, 70\% (21,000) female to 30\% (9,000) male. The primary attribute of smiling was held constantly balanced with each of the manipulated datasets containing 15,000 smiling and 15,000 non-smiling examples.

Results were then collected across each of the imbalance levels at varying levels of noise while holding all other privacy-related hyperparameters constant ($\delta=10^{-6}, S=5$). The noise multiplier, $z=\sigma S$, was varied to ensure a ratio between noise and gradient clipping was enforced. 
This led to the consideration of three different differential privacy levels across each of the imbalance datasets: high ($\epsilon=1.15, \delta=10^{-6}$, $z$ = 1.5) medium ($\epsilon=4.98, \delta=10^{-6}$, $z$ = 0.7) and low ($\epsilon=16.2, \delta=10^{-6}$, $z$ = 0.5). A baseline without any differential privacy applied (Non-DP) was also recorded. 

Furthermore, gradient clipping was varied ($S=1, S=5, S=10$) and studied using the 70\% female to 30\% male dataset, across the three different privacy levels.  

\subsection{Impact of DP on Accuracy at Test Time}
Empirically we find that model utility has an inverse relationship with information leakage- as greater privacy guarantees are placed on the model, the overall accuracy of the model decreases. An example of our results are shown in Table 1. This observation agrees with the work of \citet{alvim2011differential}. 

\begin{table}[]
\caption{Overall accuracy results and the delta when compared to the non-DP baseline for the 90\% female imbalanced dataset.}
\label{tab:my-table}
\begin{tabular}{lll}
\hline
\textbf{Privacy level}   & \textbf{Accuracy (\%)} & \textbf{$\Delta$ (\%)} \\ \hline
High ($\epsilon=1.15$)   & 71.05                  & -20.6                \\ \hline
Medium ($\epsilon=4.98$) & 85.61                  & -6.04                \\ \hline
Low ($\epsilon=16.2$)    & 87.51                  & -4.14                \\ \hline
Non-DP                   & 91.65                  & 0                  \\ \hline
\end{tabular}
\end{table}

\begin{figure}
    \centering
     \includegraphics[width=0.85\linewidth]{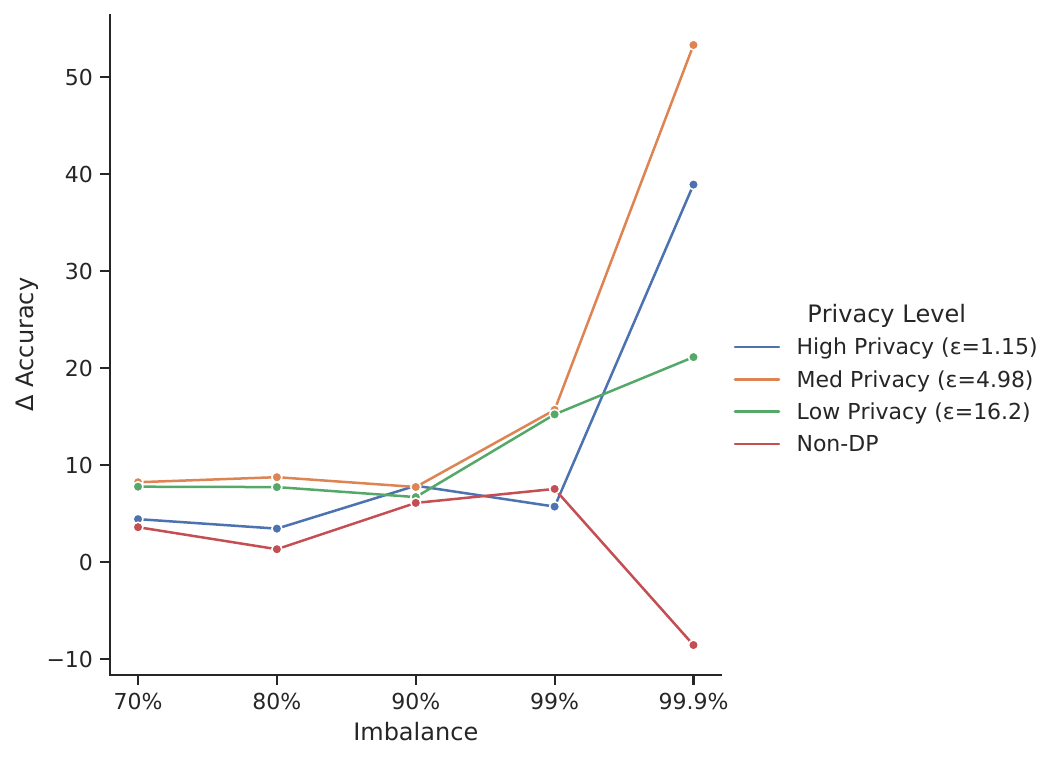}
    \vspace{-2ex}
    \caption{Smile detection accuracy delta between subgroups (female - male) vs. dataset imbalance at varying levels of DP.}
    \label{fig:acc}
    \vspace{-1ex}
\end{figure}

\begin{figure}
    \centering
     \includegraphics[width=0.85\linewidth]{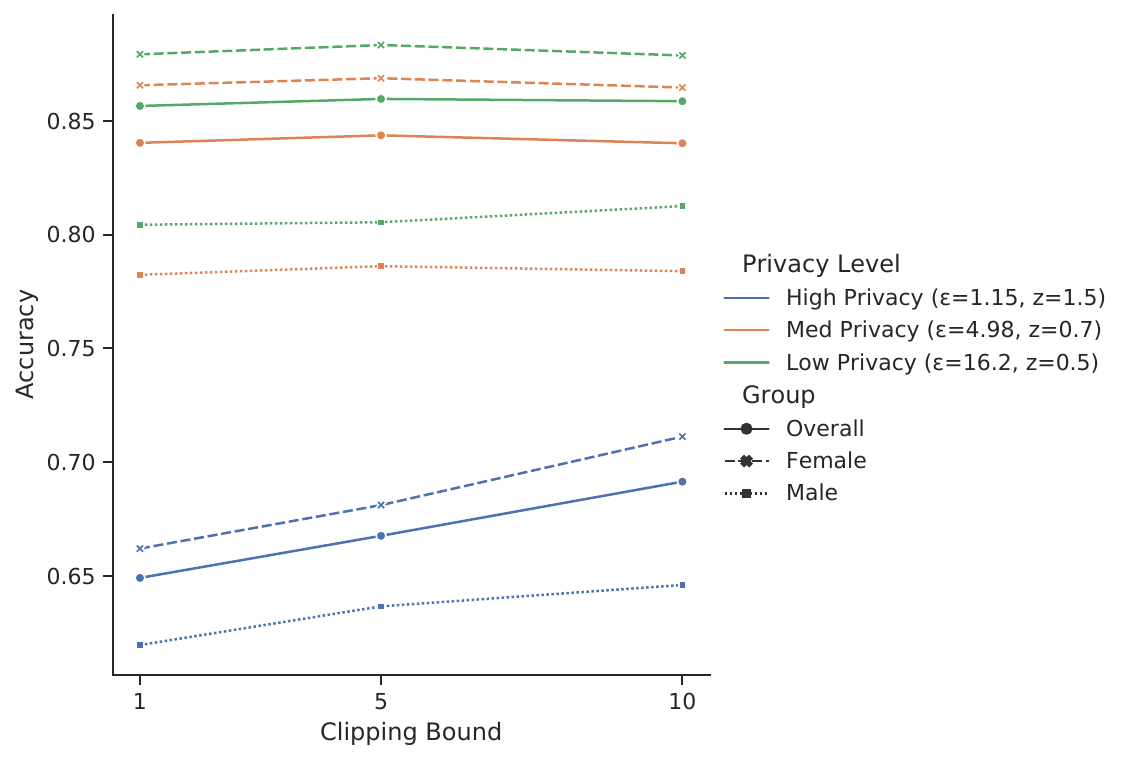}
    \vspace{-2ex}
    \caption{Smile detection accuracy across groups vs. gradient clipping at varying levels of DP.}
    \label{fig:acc2}
    \vspace{-1ex}
\end{figure}

Figure~\ref{fig:acc} shows that data imbalance appears to have little effect on the difference ($\Delta$) between subgroup accuracies (i.e. female accuracy minus male accuracy) until the split between the subgroups is extreme at 99.9\% vs 0.1\%. This is demonstrated by the similar accuracy deltas across all privacy levels up until the most extreme level of data imbalance at which point accuracy deltas diverge. Interestingly, up to the most extreme level of imbalance, it is the highest privacy level ($\epsilon=1.15$) which most closely matches the non-DP baseline. We hypothesise that at this level of differential privacy the noise and clipping techniques applied during the training process obfuscate the gradient signal to such an extent whereby all samples have a similar training effect. Therefore, while the overall model utility reduces, due to noisier gradients, the utility difference between subgroups actually improves. This phenomenon is seen again in experiments with different clipping levels, as shown in Figure~\ref{fig:acc2}, where the strictest privacy guarantee has the smallest accuracy difference between subgroups (4.24\%). This is an important takeaway; increasing the privacy does not necessarily make the utility disparity worse. There is a point in which the utility is diminished so much, that the classifier is almost random, and at that point, it actually becomes less disparate.

\subsection{Impact of DP on Accuracy During Training}

\begin{figure}
    \centering
     \includegraphics[width=0.8\linewidth]{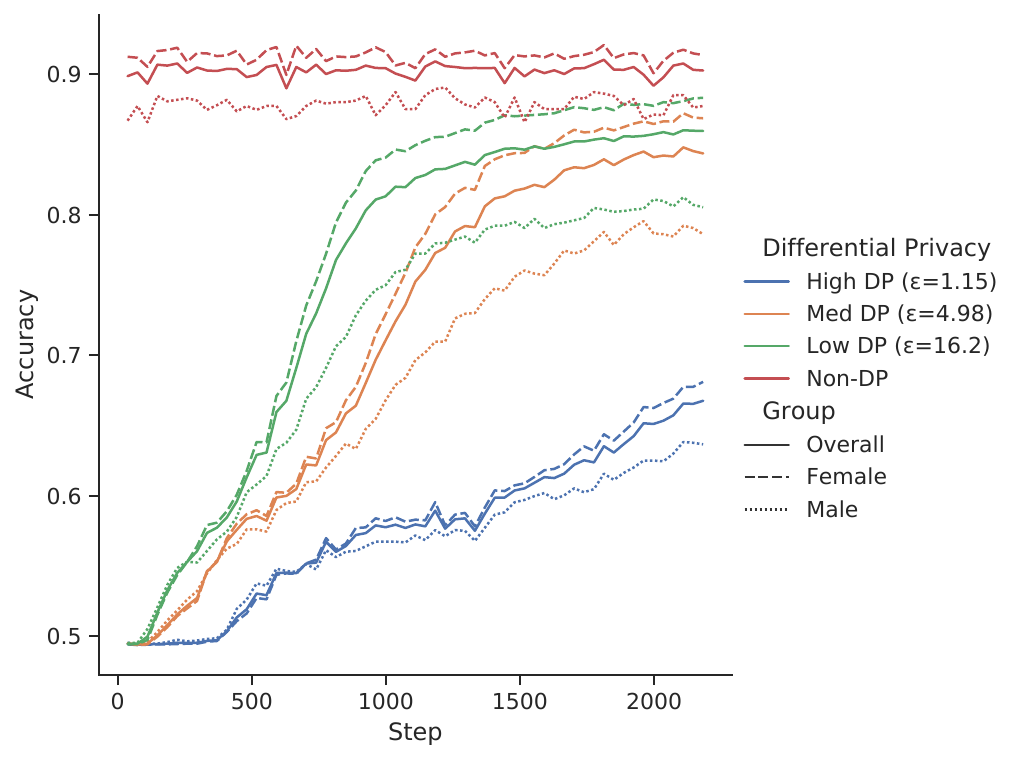}
    \vspace{-2ex}
    \caption{Smile detection accuracy for differing groups (overall, female, male) vs. training step at varying levels of DP on the 70\% female dataset.}
    \label{fig:acc3}
    \vspace{-1ex}
\end{figure}

From Figure~\ref{fig:acc3} it is clear that the training accuracy is highest without any privacy guarantee (Non-DP) and the accuracy across all groups (overall, female, male) at all time steps reduces as stricter privacy guarantees are used. We see that even very weak privacy guarantees ($\epsilon=16.2$) significantly impact the rate of convergence in training, as well as the final utility of the trained model. 

Moreover, Figure 3 shows that when conducting training efforts without any privacy guarantees (Non-DP) the difference in accuracy disparity ($\Delta$ accuracy) between female and male subgroups remains relatively constant. However, for all differentially private training runs the accuracy delta between female and male subgroups diverges throughout training, with lower levels of privacy ($\epsilon=4.98, \epsilon=16.2$) suffering from greater divergence than stricter measures. 

We hypothesise that the divergence observed during training with DP-SGD is due to the gradient clipping which bounds the influence of outliers. There are fewer examples of the minority group in each training mini-batch thus their gradients are higher and therefore more likely to be clipped. This hampers the model's ability to learn effectively from the minority group meaning that the bulk of the accuracy gain of the model is obtained from performance increases on the majority group. This effect then compounds over the course of model training; as the model becomes more proficient at accurately detecting the majority group, an increasing proportion of the minority group training examples become outliers, with clipped gradients, leading to a diverging accuracy gap between the subgroups. 

\begin{figure}
    \centering
     \includegraphics[width=0.8\linewidth]{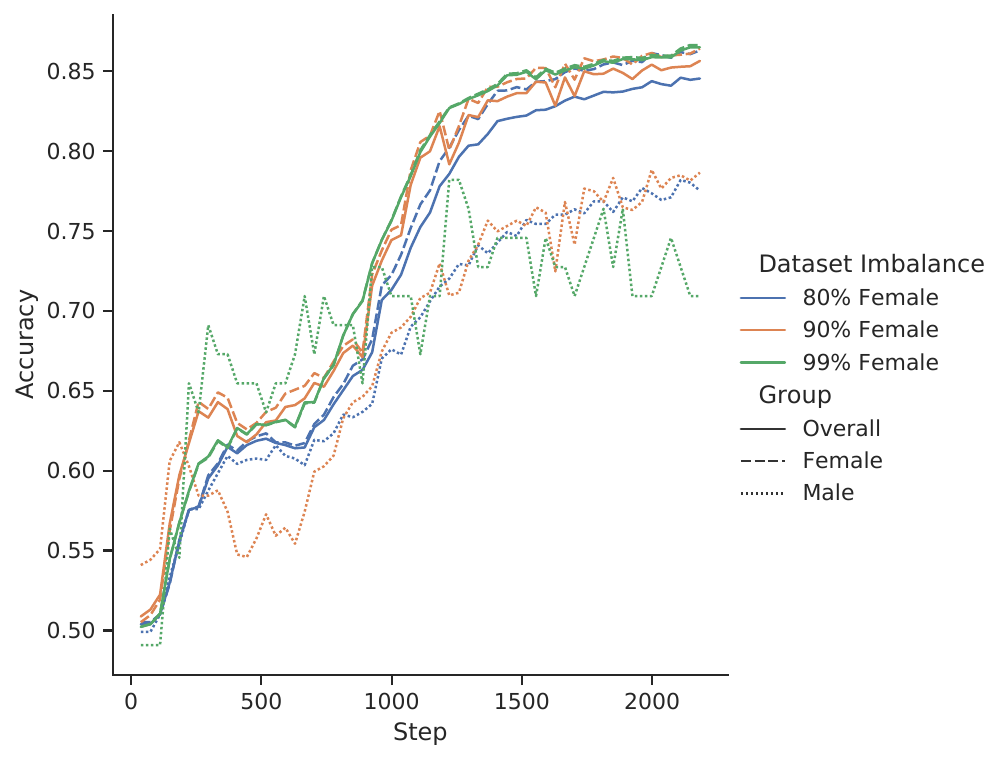}
    \vspace{-2ex}
    \caption{Smile detection accuracy for differing groups (overall, female, male) vs. training step at varying levels of dataset imbalance. Trained with ($\epsilon=4.98, \delta=10^{-6}$)-DP.}
    \label{fig:acc4}
    \vspace{-1ex}
\end{figure}

Figure~\ref{fig:acc4} demonstrates that more imbalanced datasets cause a larger divergence in accuracy between subgroups during the course of training. This reinforces the above hypothesis; datasets with smaller minority groups are more quickly impacted by the effects of gradient clipping leading to more glaring divergence over time. 

\subsection{Impact of DP on Fairness Metrics}

\begin{figure}
    \centering
        \begin{subfigure}{0.5\textwidth}
         \includegraphics[width=0.7\linewidth]{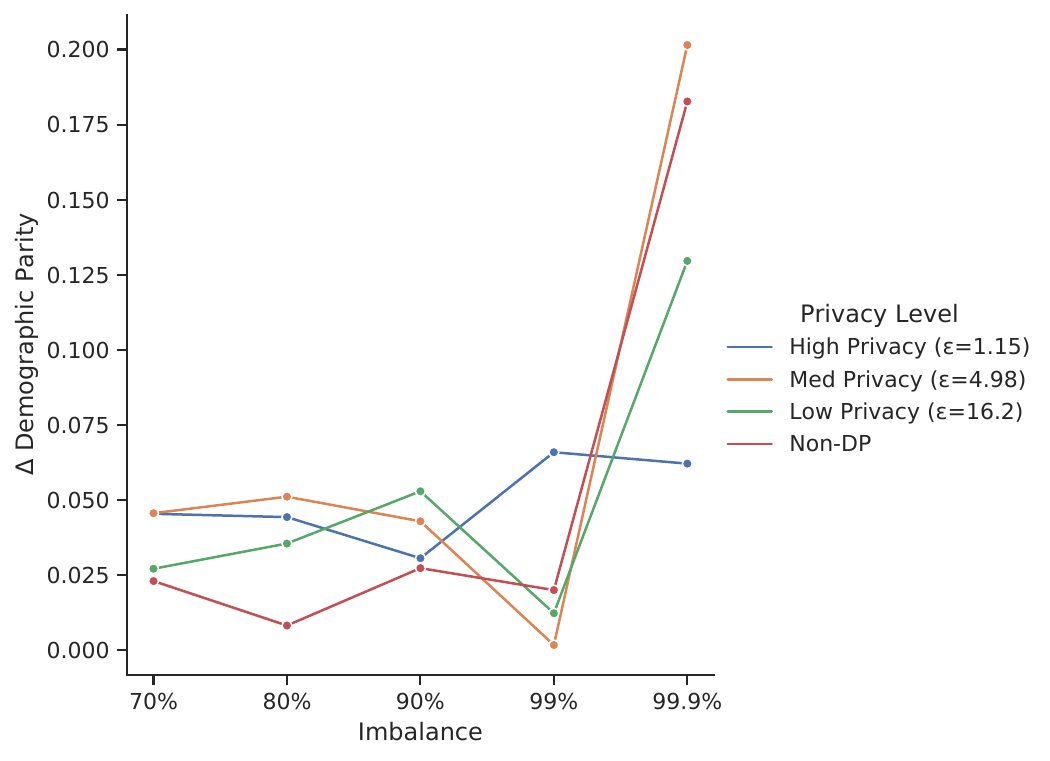}
     \caption{$\Delta_{DemP}$}
     \label{fig:time:female}
    \end{subfigure}
    \begin{subfigure}{0.5\textwidth}
     \includegraphics[width=0.7\linewidth]{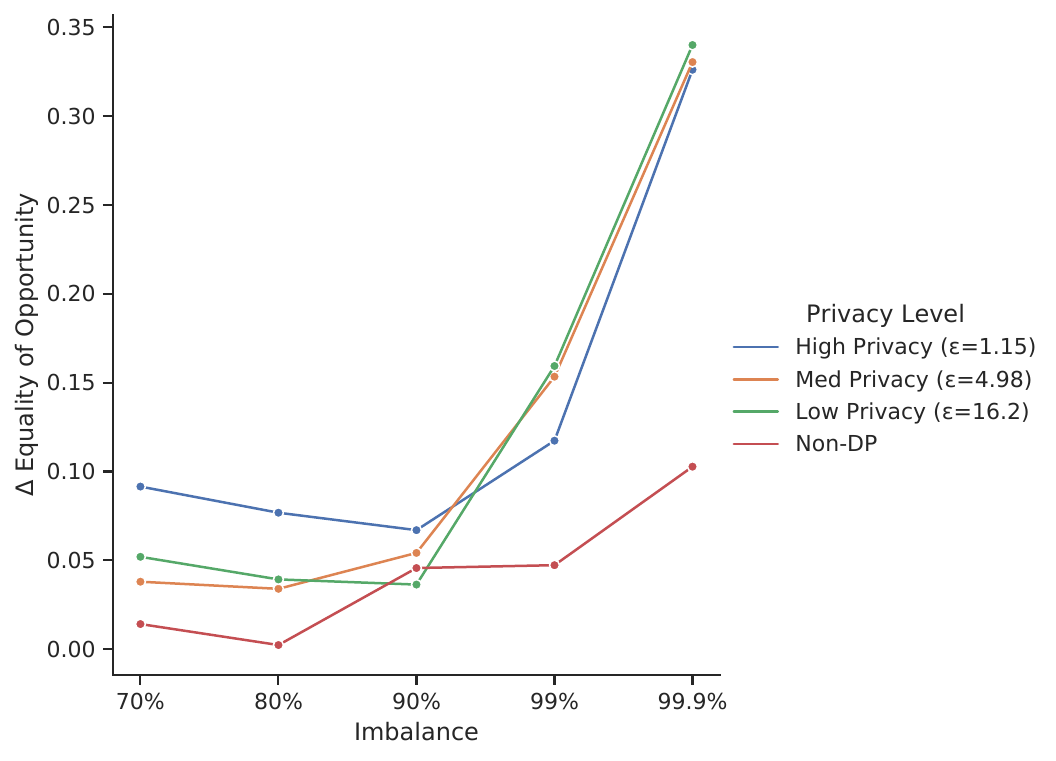}
     \caption{$\Delta_{EO}$}
     \label{fig:time:male}
    \end{subfigure}
    \vspace{-2ex}
    \caption{(a) $\Delta_{DemP}$ and (b) $\Delta_{EO}$ vs. dataset imbalance at varying levels of DP.}
    \label{fig:fair1}
    \vspace{-1ex}
\end{figure}

To measure the effects of DP on fairness, we have chosen two metrics that are commonly used in the literature: difference in Demographic Parity ($\Delta_{DemP}$) and the difference in Equality of Opportunity ($\Delta_{EO}$)~\cite{hardt2016equality}. Figure~\ref{fig:fair1} shows the results.
In a classification task, demographic parity requires the conditional probability of the classifier predicting output class $\hat{Y}= y$ given sensitive variable $S = 0$ to be the same as predicting class $\hat{Y} = y$ given $S=1$. Here the sensitive feature is gender. $\Delta_{DemP}$ is the difference between these two probabilities and the lower this difference, the more fair the classifier.
This metric is not suitable for the cases where the target and sensitive variables are correlated. This does not concern our case, however, since smiling and gender are unrelated.

Equality of opportunity is another fairness measure, which requires the conditional probability of the classifier predicting class $\hat{Y}= 1$ given sensitive variable $S=0$ and ground truth class $Y=1$ be equal to the same conditional probability but with $S=1$. Similar to the demographic parity case, we also measure the difference in these conditional probabilities.
The overall trend that we observe in Figure~\ref{fig:fair1} is that when we move to high levels of data imbalance, both the fairness metrics get worse, across all levels of privacy.
The trend is similar to that of Figure~\ref{fig:acc}, especially for the $\Delta_{EO}$. We can see that there are some small differences between the two fairness metrics, especially in the high privacy model, and especially for the highly imbalanced case. We assume that the fact that the high privacy model is doing better in terms of demographic parity and worse in terms of equal opportunity is due to the precision of the high privacy model actually getting worse. 

%%%%%%%

\section{Conclusion And Future Directions}
In this work, we study how different levels of imbalance in the data can have disparate effects on the model accuracy as well as the fairness of the decisions.
Our studies yield the following important results: a) The disparate impact of differential privacy on model accuracy is not limited to highly imbalanced data and can occur in situations where the classes are slightly imbalanced. b) The disparate impacts are not limited to high privacy levels- even for loose guarantees, DP has disparate impacts on model accuracy. c) By increasing the privacy level, we do not always see an increase in disparate impacts, since tighter privacy guarantees degrade the utility so much that the model becomes more random and therefore more fair. 
For future work, we plan to delve deeper and explore the effects of imbalanced data for tasks related to natural language processing. Furthermore, we wish to propose a mitigation mechanism that would help discover bias and decrease it, using recourse or semi-supervised learning. We encourage future research direction in this space with different data modalities. 
%

% Acknowledgements should only appear in the accepted version.

% In the unusual situation where you want a paper to appear in the
% references without citing it in the main text, use \nocite
% \nocite{langley00}

\bibliography{sample-base}
\bibliographystyle{ACM-Reference-Format}

\end{document}